\documentclass[a4paper,fleqn]{cas-dc}
\usepackage[numbers]{natbib}
\usepackage{algorithm} 
\usepackage{algpseudocode} 
\usepackage{listings}
\usepackage[
singlelinecheck=false % <-- important
]{caption}

\usepackage{array}
\usepackage{multirow}
\usepackage{tabularx, booktabs}
\usepackage{amsmath}
\usepackage{amssymb}

\newcolumntype{L}[1]{>{\raggedright\let\newline\\\arraybackslash\hspace{0pt}}m{#1}}
\newcolumntype{C}[1]{>{\centering\let\newline\\\arraybackslash\hspace{0pt}}m{#1}}
\newcolumntype{R}[1]{>{\raggedleft\let\newline\\\arraybackslash\hspace{0pt}}m{#1}}

\def\tsc#1{\csdef{#1}{\textsc{\lowercase{#1}}\xspace}}
\tsc{WGM}
\tsc{QE}
\begin{document}
\let\WriteBookmarks\relax
\def\floatpagepagefraction{1}
\def\textpagefraction{.001}

\shorttitle{Evaluating Generative Patent Language Models}    

\shortauthors{Jieh-Sheng Lee}  

\title [mode = title]{Evaluating Generative Patent Language Models}  
%\title{Evaluating Generative Patent Language Models}



\author[]{Jieh-Sheng Lee}[orcid=0000-0002-0990-6170]

%\cormark[(corr mark no)]

%\fnmark[(footnote mark no)]

\ead{jasonlee@nycu.edu.tw}

%\ead[url]{(URL)}

%\credit{Credit authorship details}

\affiliation[1]{organization={National Yang Ming Chiao Tung University School of Law},
            addressline={No. 1001, Daxue Rd. }, 
            city={Hsinchu},
%          citysep={}, % Uncomment if no comma needed between city and postcode
            postcode={300093}, 
            state={},
            country={Taiwan}}

% % Corresponding author text
% \cortext[1]{Corresponding author}

% % Footnote text


\begin{abstract}
Generative language models are promising for assisting human writing in various domains. This manuscript aims to build generative language models in the patent domain and evaluate model performance from a human-centric perspective. The perspective is to measure the ratio of keystrokes that can be saved by autocompletion based on generative patent language models. A higher ratio means a more effective model which can save more keystrokes. This metric can be used to benchmark model performance. The metric is different from conventional machine-centric metrics that are token-based instead of keystroke-based. 
%The performance of models in different sizes can also be evaluated in such a metric by measuring a number of newly granted patents. 
In terms of model size, the largest model built in this manuscript is 6B, which is state-of-the-art in the patent domain.
Based on the metric, it is found that the largest model is not necessarily the best for the human-centric metric. The finding means that keeping increasing model sizes in the patent domain might be unnecessary if the purpose is to assist human writing with autocompletion. 
Several patent language models are pre-trained from scratch in this research. The pre-trained models are released for future researchers.  
Several visualization tools are also provided.
The importance of building a generative language model in the patent domain is the potential to facilitate creativity and innovations in the future.

%This is abstract. The abstract must be between 150--250 words. 
\end{abstract}


\maketitle

% Main text
\section{Introduction}
\label{section:introduction}
Generative language models have shown great potential in recent years. To train a generative model, it is common to collect a large amount of general or diverse data and train a model to generate similar data. 
The mainstream research is to continue to build larger models with more data. The evaluation of generative language models is also machine-centric and token-based. 
In this manuscript, instead of pursuing a larger and general language model, the objective is to build a domain-specific language model and evaluate the model with a human-centric metric. The metric is to measure the ratio of keystrokes that can be saved by autocompletion based on the generative model.
When building the state-of-the-art generative patent language model, this manuscript shows that the largest model is not necessarily the best for the metric.

The metric is a post hoc analysis. A generative patent language model is a model capable of generating patent text. From a post hoc perspective, given some patent text written by a patent professional, the proposed metric can be utilized to show how close the model may generate the actual patent text.
The proximity between the generated text and the actual text can be measured by the number of keystrokes that can be saved for a user if an autocomplete function is available. The model is more effective and closer to the user's writing if more keystrokes are saved.
The long-term goal for the future is to assist patent professionals and inventors save as many keystrokes as possible. The metric proposed in this manuscript is one way to measure progress toward the goal.
In addition, the performance of the models in different sizes can be evaluated based on the metric. New patents are granted frequently and are available to the public for free. 
By measuring a number of patents collectively with the metric, it is possible to measure which model performs best.

The major contributions of this manuscript include: (1) pre-trained several language models specific to the patent domain, (2) proposed a human-centric metric for model evaluation (how many keystrokes to be saved with an autocomplete function), (3) found that the largest model is not necessarily the best, and (4) provided the tools for visualizing the experiment results.
The rest of the manuscript is organized as follows. Section~\ref{section:related_work} introduces the code base that inspires this work and previous related research. Section~\ref{section:implementation} describes the implementation of this work in detail, such as objectives, metrics, data, and model sizes. Section~\ref{section:experiments} provides experiments and visualization of the results. Section~\ref{section:implications} raises several legal implications to be addressed by legal scholars in the future, because the metric might be relevant to the nonobviousness requirement in patent law. Section~\ref{section:conclusion} concludes this manuscript.

\section{Related Work}
\label{section:related_work}

This research is inspired by works such as GPT-J-6B~\cite{gpt-j-github} and GPT-3~\cite{GPT-3-NEURIPS2020}. GPT-3 (Generative Pre-trained Transformer 3) is an autoregressive language model that uses deep learning to produce human-like text. GPT-J-6B is a transformer model using Mesh Transformer JAX, while "6B" (6 billions) represents the number of trainable parameters. 
The patent language models in this manuscript are called PatentGPT-J models because their configurations are based on GPT-J-6B's. In addition to building larger models, this manuscript also extends previous research on patent text generation. Previous research includes: generating patent claims by fine-tuning GPT-2~\cite{jiehsheng03}, controlling patent text generation by structural metadata~\cite{jiehsheng05}, and prior art search for generated patent text~\cite{jiehsheng09}. GPT-2 (Generative Pre-trained Transformer 2) is open-sourced and created by OpenAI in February 2019.
In recent years, natural language generation (NLG) has demonstrated significant progress in many domains. However, due to the entry barrier to legal knowledge, very few researchers devote themselves to the generation of patent language. This manuscript continues the research on the patent domain and aims to scale up the generative models. Unlike GPT-3, the source code and the GPT-J-6B model configuration~\cite{gpt-j-github} are available. Furthermore, in GPT-J-6B~\cite{gpt-j-github}, the training data include a portion of the patent text, and the size of the model is significantly larger than the largest GPT-2 model. Therefore, GPT-J-6B is a suitable choice for this manuscript to follow. At the time of implementation in this research, GPT-J-6B is also the largest model open-sourced. 

According to The Pile~\cite{pile}, the training data for GPT-J-6B~\cite{gpt-j-github} is an 825 GiB English text corpus targeted at training large-scale language models. The Pile is constructed from 22 diverse and high-quality datasets. Among them, the ``USPTO Backgrounds'' is a dataset of the background sections extracted from patents granted by the United States Patent and Trademark Office (USPTO). The raw data are available in the bulk archive of the USPTO website~\cite{uspto_bulkdata}. In~\cite{pile}, the authors explain that a typical patent background lays out the general context of the invention and gives an overview of the technical field. They included USPTO Backgrounds because it contains a large volume of technical writing aimed at a non-technical audience. The raw size of the dataset is 22.90 GiB. Furthermore, according to~\cite{gpt-j-github} and~\cite{gpt-j-6b-blog}, the GPT-J-6B model was trained on 400B tokens from The Pile dataset and the training took roughly 5 weeks with TPU v3-256.

\section{Implementation}
\label{section:implementation}

\subsection{Objectives}
\label{subsection:objectives}

Most of the source code in this research is forked from~\cite{gpt-j-github}. In the code base~\cite{gpt-j-github}, the GPT-J-6B model is released as the largest pre-trained model open to the public at the time. The code base also provides a guide for fine-tuning\cite{gpt-j-how-to-fine-tune} the GPT-J-6B model. The first objective of this research is to build the same largest model dedicated to the patent domain, that is, the PatentGPT-J-6B model. Instead of fine-tuning, the author pre-trained the PatentGPT-J-6B model from scratch with US patents. A corpus like patents is a specialized domain. The reason why pre-training from scratch is that it is critical for the model to generate nothing but patent text. Fine-tuning a model pre-trained from other corpora may generate non-patent text. Non-patent text makes little or no sense to patent professionals. With respect to the general requirements in training data, the patent corpus is also suitable for pre-training because the corpus is sizable and is incremental on a regular basis. 

The second objective is to measure the performance of PatentGPT-J models from a human-centric perspective. Specifically, the metric is to measure how many keystrokes the model can save for a user by providing model predictions in an autocomplete function. A language model is a statistical model that assigns probabilities to tokens after tokenizing words and sentences. The performance of language models is measured from a machine-centric perspective conventionally. For example, perplexity is a popular evaluation metric for language models. Perplexity can be defined as the exponential of the cross-entropy. In the training stage, the cross-entropy loss function encourages the model to assign high probabilities to the observed tokens. The lower the loss, the better. However, from the perspective of assisted writing by autocompletion, what loss will be sufficiently low enough? This is a question that the cross-entropy loss function cannot answer. The question has to be addressed from a different perspective. In this manuscript, the human-centric metric is used to evaluate various PatentGPT-J models of different sizes. The novel metric challenges the assumption that larger model sizes are always better.  

The third and auxiliary objective is to explore the capability of neural models by testing whether the patent-specific model may perform a general NLP task in few-shot learning. The content of the general task is not within the patent domain. If the test result is positive, what size is required for the model to demonstrate the capability of few-shot learning? Is the largest model the best?

\subsection{Metric: Autocomplete Effectiveness}
\label{subsection:metric}

In this manuscript, the metric for evaluating model performance is the Autocomplete Effectiveness (AE) ratio. The AE ratio is about how many keystrokes can be saved by using an autocomplete function. The AE ratio is defined below. A higher AE ratio is better because more keystrokes are saved.

\begin{equation}
\label{eq:1}
AE = \frac{\text{\footnotesize{(keystrokes w/o autocomplete) - (keystrokes w/ autocomplete)}}}{\text{\footnotesize{keystrokes w/o autocomplete}}}
\end{equation}

The AE ratio works at the token level. The details of implementing the AE ratio are provided below. Given a sequence of tokens \emph{$t_1$}, \emph{$t_2$}, ..., \emph{$t_k$}, the numerator of the ratio is the accumulated number of keystrokes that can be saved by an autocomplete function at each token position. For each token \emph{$t_i$}, its rank in model prediction determines whether the autocomplete function will be triggered. In programmer's popular editors for code completion, it is common to have an autocomplete function. For example, a user can press a special key, such as the ``tab'' key, to complete the next function name, parameter, or the remaining of the statement in the source code. In this manuscript, the autocomplete function works similarly, except that the target to complete is the next token in the patent text.
Without involving the actual implementation of the autocomplete function, the calculation of the AE ratio can be explained later in pseudocode. The calculation is implemented based on the predictions and ranks of model inference in an emulated fashion. In the following experiments, no actual autocomplete function of an editor is required. 
In the past, to demonstrate how fluent and coherent the generated patent text can be, an autocomplete prototype was implemented to complete an entire patent claim as a proof of concept in~\cite{jiehsheng03} based on GPT-2~\cite{gpt2_Radrof01}.  

In an emulated fashion, the calculation of the total number of keystrokes is based on how a user may compose patent text with minimal number of keystrokes in an editor. For example, given some patent text, if the editor has no autocomplete function, the minimum number of keystrokes for a user to compose the patent text is the total number of characters of the entire patent text (ignoring special function keys such as ``shift'', ``CapsLock'' and ``tab''). If the editor provides an autocomplete function, the total number of keystrokes is likely to decrease. The user can press the conventional ``tab'' key to autocomplete the next token and type less. The performance of the model can be evaluated by the degree to which the user can type fewer words. Therefore, a model assisting the user with fewer keystrokes is a better model.  

New patents are granted on a frequent basis. These new patents are free and suitable for evaluating the AE ratio in this research. In the emulated calculation, the cut-off of ranks in model prediction is set as 10.
Given a sequence of tokens \emph{$t_1$}, \emph{$t_2$}, ..., and \emph{$t_k$}, 
the \emph{rank} of the token \emph{$t_k$} is its position in the model's predictions based on tokens \emph{$t_1$}, \emph{$t_2$}, ..., and \emph{$t_{k-1}$}. The actual \emph{$t_k$} may or may not locate within the \emph{top-10} tokens predicted by the model.  
In this manuscript, the hypothetical autocomplete function works by showing the \emph{top-10} tokens predicted by the model, based on tokens \emph{$t_1$}, \emph{$t_2$}, ..., \emph{$t_{k-1}$}. 
If the \emph{top-1} token is exactly the token \emph{$t_k$}, it means that a user can press the ``tab'' key to complete the next token. If the \emph{top-2} token is the token \emph{$t_k$}, the user can press ``downarrow'' ($\downarrow$) key and ``tab'' key to complete. If the token \emph{$t_k$} is not within the \emph{top-10}, it means that the user would have to type manually without the autocomplete function. It is assumed that the length of ten is effective for a glance and that a longer list can be unproductive. 
In brief, depending on whether the token \emph{$t_k$} is within the \emph{top-10} tokens predicted by the model, the keystrokes may come from the autocomplete function or manual typing. In such settings, the number of keystrokes is accumulated by counting the ``tab'', the ``downarrow'' ($\downarrow$) key, and the number of characters in the user's manual typing.

The pseudocode to calculate the minimum number of autocomplete keystrokes is provided in Algorithm~\ref{alg:counting_keystrokes}. As described, an emulated user can confirm the \emph{top-1} token by pressing the ``tab'' key to complete. The ``tab'' counts as one keystroke. If the token \emph{$t_k$} is the token with rank as two in model prediction, the autocomplete function will require pressing the ``downarrow'' key and the ``tab'' key to select the target token. Two keystrokes will be counted accordingly.  
If the token \emph{$t_k$} is the token with rank as three in model prediction, using the autocomplete function will require three keystrokes (two ``downarrow'' keys and one ``tab'' key). As a general rule, the number of keystrokes is equal to the rank matched shown in the \emph{top-10} list predicted by the model. By emulating how the sequence of tokens \emph{$t_1$}, \emph{$t_2$}, ..., \emph{$t_k$} may be composed with the hypothetical autocomplete function, it is feasible to count the total number of keystrokes to complete the entire text of the token sequence.

%\hfill
\begin{algorithm}
  \caption{Calculating Autocomplete Keystrokes}\label{alg:counting_keystrokes} 
  \begin{algorithmic}[1]
    \State {$patent\_text$} $\leftarrow$ read one new patent claim
    \State {$patent\_tokens$} $\leftarrow$ tokenizer.encode({$patent\_text$})  
    \State $keystroke$ $\leftarrow$ 0
    \For {$i=1$ to len($patent\_tokens$)-1}
      %\State {$prompt\_tokens$} $\leftarrow$ $patent\_tokens$[:i]
      \State {$prompt$} $\leftarrow$ $patent\_tokens[:i]$
      \State {$next\_token$} $\leftarrow$ $patent\_tokens[i]$
      %\State {$rank$} $\leftarrow$ $rank(next\_token)$
      \State $top$ $10$ $tokens$ $\leftarrow$ model.predict($prompt$)
      %\State display $top\_10\_tokens$ at cursor position
      % \State $selection$ $\leftarrow$ $top\_1\_token
      \If {$next\_token$ == $top$ $1$ $token$ }
        \State $keystroke$ += 1 \Comment{autocomplete by ``tab'' key}
      \ElsIf {$next\_token$ == $rank$ $n$ $token$} \Comment{$n$ = $2$ to $10$}
        \State $next\_text$ $\leftarrow$ tokenizer.decode({$next\_token$})  
        \State $keystroke$ += min($n$, len($next\_text$)) \Comment{auto or manual}
      \ElsIf {$next\_token$ is out of $top$ $10$}
      %\Else \Comment{$next\_token$ is out of $top$ $10$}
        \State $keystroke$ += len($next\_text$) \Comment{manual typing}
      \EndIf
    \EndFor

  \end{algorithmic} 
\end{algorithm}

A caveat is that if the text length of the matching token in the \emph{top-10} list is shorter than the number of keystrokes by pressing the ``tab'' and the ``downarrow'', it will be more efficient for a user to type the token text manually. Therefore, in this scenario, the number of keystrokes should be the text length of the match token due to fewer keystrokes. In a nutshell, if the token \emph{$t_k$} matches the \emph{top-1} token in model prediction, the keystroke counts as one. If the token \emph{$t_k$} matches a token in model prediction (\emph{rank = 2$\sim$10}), the keystroke(s) will count as \emph{n} or the length of the token text, which is smaller. If the rank of the token \emph{$t_k$} is beyond 10 and not shown in the autocomplete function, a user will have to type the token text manually. The number of keystrokes therefore equals to the text length of the token \emph{$t_k$}.   

It should be noted that MRR (Mean Reciprocal Rank) is a common metric for measuring model performance. For example, in~\cite{CodeFill} and~\cite{Learning_Autocompletion}, the authors use the metric to measure code completion and describe: MRR assesses the whole top \emph{N} recommended completions and takes into account the first position in which the target is matched. For a single query, the reciprocal rank is $\frac{1}{rank}$ where \emph{rank} is the position of the highest-ranked answer. If no correct answer was returned in the query, then the reciprocal rank is 0. Compared to the AE ratio in this manuscript, the reciprocal rank serves a similar purpose in measuring autocomplete effectiveness. However, a non-zero MRR value does not precisely reflect the actual human effort in terms of keystrokes. The number of keystrokes is also lost if the MRR value is zero. These are the reasons why this research takes a different metric from a more human-centric perspective. Furthermore, an MRR metric aggregates the calculations of reciprocal ranks on average. Without any other visualization, the distribution of ranks for each answer/token is lost in the mean value of the MRR. In this research a visualization is provided to inspect the distributions of ranks in this manuscript.

\subsection{Data}
\label{subsection:data}
A granted patent has several sections in the patent document. This manuscript is motivated to have a broader coverage and complement the USPTO Backgrounds dataset in The Pile. The dataset in this research is named the USPTO TACD dataset to include titles, abstracts, claims, and descriptions. Both independent and dependent claims are included. The raw data ranges from 1976 to 2021 (1976$\sim$2020 for training and 2021 for validation). For a benchmark purpose, the USPTO TACD dataset does not include the background section. Interested researchers may combine the USPTO TACD dataset and the USPTO Backgrounds dataset, if necessary. The script to collect the USPTO Backgrounds dataset is provided in the repository~\cite{EleutherAI_pile_uspto}, which takes advantage of the code in~\cite{cfoster0_uspto-patent-data-parser} (forked from~\cite{TamerKhraisha_uspto-patent-data-parser}). The author of this manuscript leveraged these repositories to build the USPTO TACD dataset. The downloaded and shuffled patent text for building the USPTO TACD dataset is compressed and available upon request.

As mentioned in The Pile~\cite{pile}, the background section of a patent document is aimed at a non-technical audience. This should be the reason why the USPTO Backgrounds dataset was built and included in The Pile. On the contrary, the USPTO TACD dataset in this manuscript aims at a more technical audience. Especially, patent claims aim at patent professionals. From a legal perspective, patent claims are the most important part of a patent document because they define the scope of the legal protection conferred by a patent. Among patent claims, a unique feature is claim dependency: A patent claim may depend on another patent claim. A patent claim without any dependency is called an independent claim. A patent claim that depends on another patent claim is called a dependent claim. Compared to other corpus in plain text, the claim dependency poses a unique challenge and opportunity in capturing its structure. The approach proposed in this research to expand the text of patent claims will be explained in~\ref{subsection:claim_expansion}. The approach makes the USPTO TACD dataset significantly larger than the downloaded raw text and the USPTO Backgrounds dataset. 

Another factor enlarging the USPTO TACD dataset is related to bidirectional text generation. The patent text is a source of inventive ideas. It is hypothesized that a generative patent language model can generate new patent text for inventor brainstorming. In the use case of brainstorming, it is not necessary that an inventor starts from the beginning of patent text. The inventor may start from the middle or from anywhere. Therefore, it is preferable that the generative language model can generate patent text bidirectionally. Therefore, the patent text is duplicated in reverse order as part of the dataset.  

After adding data augmentation with claim dependency and reversed order, the total size of the USPTO TACD dataset in text is about 713 GiB. The size is significantly larger than the 22.90 GiB of the USPTO Backgrounds dataset. In comparison, the size of The Pile is 825 GiB. In terms of tokens, the total number of the USPTO TACD dataset is about 147B, compared to the 400B tokens in The Pile. The tokens required for encoding one GiB of raw text are fewer for PatentGPT-J models because the tokenizer is pre-trained from scratch with patent text, too. By using the pre-trained tokenizer, the efficiency for encoding the USPTO TACD dataset in tokens is higher. In this research, pre-training a PatentGPT-J model consumes about 11B tokens for training 350,000 steps. The 11B tokens are about 7.48\% of the USPTO TACD dataset (147B). It is noted that the vocabulary for the pre-trained tokenizer contains some special tags defining the unique structure and meta data in patent documents. The practice of using domain-specific tags, such as \texttt{<|start\_of\_claim|>}, \texttt{<|dep|>}, and \texttt{<|end\_of\_claim|>}, was implemented in previous research~\cite{jiehsheng05} and reused.

\subsection{Model sizes \& Training Loss}
\label{subsection:model_size}

The pre-training from scratch is done following the fine-tuning guide~\cite{gpt-j-how-to-fine-tune} and without using any saved checkpoint. 
Based on the code base~\cite{gpt-j-github}, the configuration of the GPT-J-6B model includes 4,096 dimensions (\emph{d\_model}) and 28 layers (\emph{layers}). The model size of the GPT-J-6B is 6B. In this research, by changing the number of dimensions, layers, or both, several PatentGPT-J models are pre-trained and configured in different sizes. The model sizes in this research are 6B, 1.6B, 456M, 279M, 191M, 128M, and 115M, as shown in Table~\ref{table:model_comparison}. It is noted that the original GPT-J-6B was pre-trained with TPU v3-256 (\emph{tpu\_size = 256}) and the training steps are 383,500. In the fine-tuning guide~\cite{gpt-j-how-to-fine-tune}, the training step is configured as 350,000. PatentGPT-J models follow the fine-tuning guide and train 350,000 steps. In terms of computing resource, this research utilizes TPU v3-8 which is sufficient for this stage of research. TPU v3-8 is more accessible and cost less. Consequently, \emph{tpu\_size} is set as 8 in the configuration.

\begin{table}[]
\caption{Model Comparison}
\label{table:model_comparison}
\centering
  \begin{tabular}{c c c c c c}
    \hline
    model & size & layer & d\_model & tpu\_size & steps\\ \hline\hline
    PatentGPT-J & 115M & 1 & 1024 & 8 & 350,000 \\ \hline
    PatentGPT-J & 128M & 2 & 1024 & 8 & 350,000 \\ \hline
    PatentGPT-J & 191M & 7 & 1024 & 8 & 350,000 \\ \hline
    PatentGPT-J & 279M & 14 & 1024 & 8 & 350,000 \\ \hline
    PatentGPT-J & 456M & 28 & 1024 & 8 & 350,000 \\ \hline
    PatentGPT-J & 1.6B & 28 & 2048 & 8 & 350,000 \\ \hline
    PatentGPT-J & 6B & 28 & 4096 & 8 & 350,000 \\ \hline\hline
    GPT-J       & 6B & 28 & 4096 & 256 & 383,500 \\ \hline
  \end{tabular}
\end{table}

\begin{figure}[ht]
  {\includegraphics[width=0.5\textwidth, keepaspectratio]{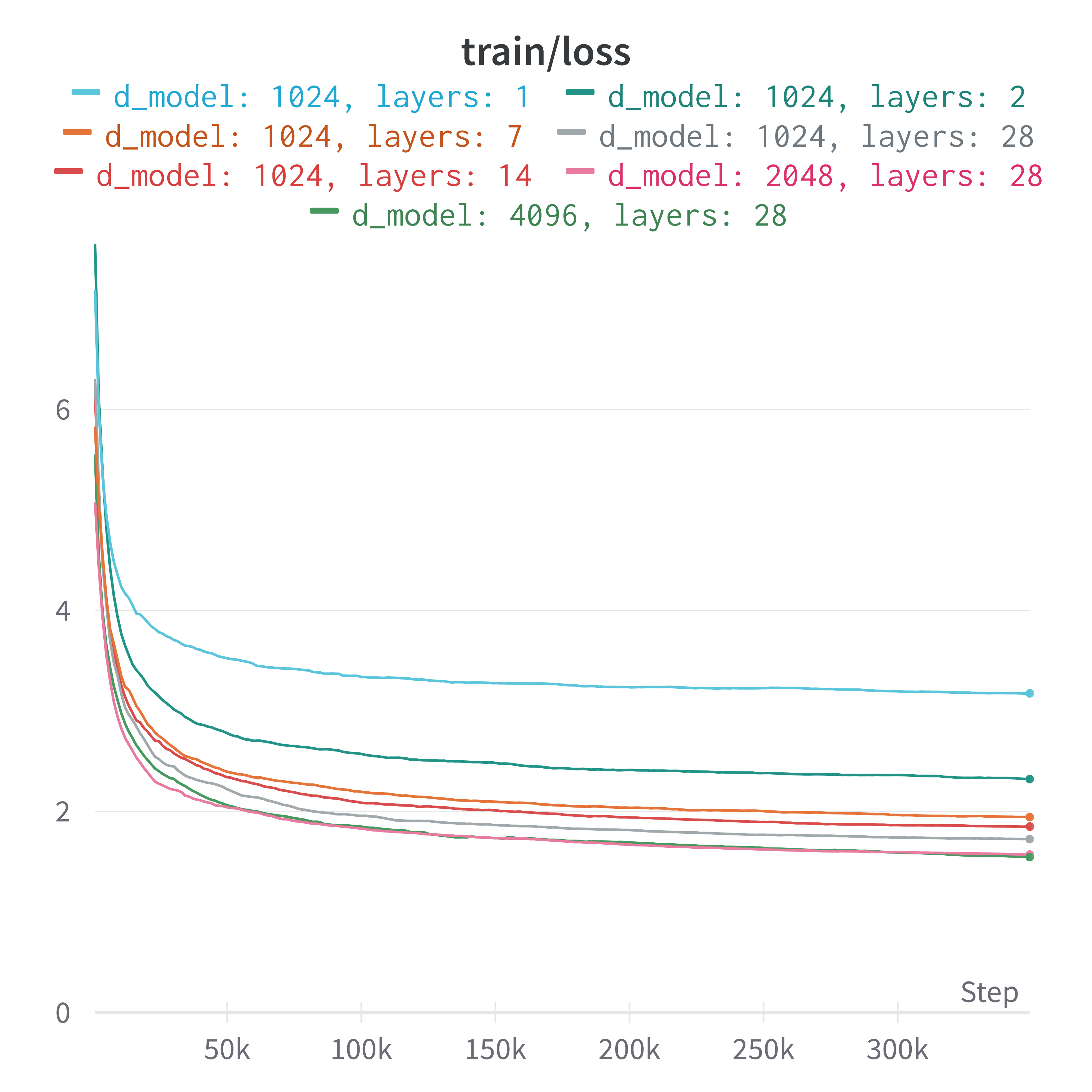}}
  \caption{Training loss curve}
  \label{train_loss}
\end{figure}

Fig.~\ref{train_loss} shows the training loss of PatentGPT-J models. In terms of training loss, both the 6B model (\emph{d\_model=4096}) and the 1.6B model (\emph{d\_model=2048 and layer=28}) have lower values ranging from 1.56 to 1.53~\cite{PatentGPT-J-training_loss} around the end of training. Compared to GPT-J-6B, its training loss ranges from 1.59 to 1.39 around the end according to~\cite{GPT-J-training_loss}.
In this investigation, it is found that the model with the lowest loss is not necessarily the best for the AE ratio. 
As shown in Section~\ref{section:experiments}, the 456M model (\emph{d\_model=1024 and layer=28}, the third curve from the bottom in Fig.~\ref{train_loss}) has the highest AE ratio in the experiment in~\ref{subsection:experiment_1}. The 1.6B model has the highest AE ratio in the experiment in~\ref{subsection:experiment_2}. The 6B model is not the best in both experiments. One probable reason is that the AE ratio takes into account the length of the token text. This length of text is not included in the calculation of the training loss. Therefore, the training loss in cross-entropy does not align with the metric squarely. For future researchers, the pre-trained PatentGPT-J models in 279M, 456M, and 1.6B, including checkpoints and vocabulary files, are released at~\cite{jiehsheng_PatentGPT-J_github}. The sample code at~\cite{jiehsheng_PatentGPT-J_github} also shows how to use a PatentGPT-J model for inference. The sample code is for the sentiment analysis in section~\ref{subsection:experiment_4}.

\subsection{Claim Dependency}
\label{subsection:claim_expansion}
This section explains how to expand the claim text based on the claim dependency. For patent professionals in daily practice, it is common for an independent claim to have several dependent claims that describe different technical elements or limitations in the claimed invention. The several dependent claims do not, in general, depend on each other. For example, \emph{claim 2}, \emph{claim 3}, ..., and \emph{claim n} are several dependent claims and depend only on \emph{claim 1}, and there is no claim dependency among \emph{claim 2}, \emph{claim 3}, ..., and \emph{claim n}. In such a scenario, a patent language model should learn only the dependency between \emph{claim 1} and \emph{claim 2}, \emph{claim 1} and \emph{claim 3}, ..., and \emph{claim 1} and \emph{claim n}. However, in the raw text downloaded from the USPTO, the claims are described in sequence such as \emph{claim 1}, \emph{claim 2}, \emph{claim 3}, ..., and \emph{claim n} consecutively. The \emph{claim x (x=2$\sim$n)} is implicitly dependent on all of it previous claim(s) including \emph{claim 1} to \emph{claim x-1} because the language model is an autoregressive model. Such an implicit claim sequence in the raw text should not be treated as a claim dependency in the dataset. 

The objective of a language model is to predict the next token based on all preceding tokens. Therefore, without removing the implicit sequence and incorrect dependency, after feeding the raw patent text, predicting \emph{claim n} will mean predicting the claim based on \emph{claim 1}, \emph{claim 2}, \emph{claim 3},... and \emph{n-1} altogether. However, as described, the claim dependency between \emph{claim n} and \emph{claim 1} is remote. \emph{Claim n} should depend on \emph{claim 1} only and nothing in between. \emph{Claim 2}, \emph{claim 3}, ..., and \emph{claim n-1} should not be involved when training \emph{claim n} and \emph{claim 1}.  
In the training dataset, the dependency in the pair (\emph{claim n} and \emph{claim 1}) should be only one record. The other pairs, such as \emph{cliam x (x=2$\sim$n)} and \emph{claim 1} should be separate training records too. 
To correctly reflect the dependency of claims, this research takes an expansive approach to mitigate the above issues.
The approach is to prepare a pair of claims as a training record according to the dependency between the claims. The pair is prepared by concatenating these two claims with a special token \emph{$<$|dep|$>$} to denote the dependency.
For example, if \emph{claim n2} depends on \emph{claim n1}, the training data will be prepared as ``\emph{claim n1$<$|dep|$>$claim n2}.'' 
The paired structure reflects the dependency of the two claims. 
In this way, ``\emph{claim 1$<$|dep|$>$claim 2}'' is a training record, and ``\emph{claim 1$<$|dep|$>$claim 3}'' is another record. The duplication of the \emph{claim 1} for all pairs ``\emph{claim 1$<$|dep|$>$claim x (x=2$\sim$n)}'' is the reason why the USPTO TACD dataset in this research is large. 

It should be noted that the expansion of claims in this manuscript is limited to paired claims only. In reality, sometimes the situation could be more complicated than the general scenario of paired claims. A dependent claim may depend on another dependent claim or depend on multiple claims (independent or dependent). For example, \emph{claim n3} may depend on \emph{claim n2}, and \emph{claim n2} may depend on \emph{claim n1}. In this situation, the claim dependency is like a chain of dependency. The USPTO TACD dataset does not formulate this chain as one single training record. It is hypothesized that the paired claims in this manuscript can make the model learn the dependency in the chain discretely in pairs. This hypothesis will be validated in the future.
Another situation not addressed in this research is multiple-dependent claims. For example, \emph{claim n6} depends on \emph{claim n4} and also on \emph{claim n5}. In this scenario, \emph{claim n6} is a multiple-dependent claim.

\section{Experiments \& Visualization} 
\label{section:experiments}

The first experiment in section~\ref{subsection:experiment_1} and the second experiment in section~\ref{subsection:experiment_2} evaluate the PatentGPT-J models based on the AE ratio defined in Equation~\ref{eq:1}. A model performs better if its AE ratio is higher. 
The third experiment in section~\ref{subsection:experiment_3} is to test the models with non-patent text. The purpose is to observe how much the AE ratio will drop. 
The fourth experiment in section~\ref{subsection:experiment_4} is sentiment analysis. During this research, it was found by accident that larger PatentGPT-J models can perform sentiment analysis reasonably well in few-shot learning. The content of the sentiment analysis is outside the patent domain. PatentGPT-J models are trained specifically for the patent domain only. Therefore, a patent language model capable of doing sentiment analysis is counterintuitive. To visualize the results of the experiment, statistics and figures are provided in section~\ref{subsection:statistics} and an inspection tool is demonstrated in section~\ref{subsection:interactive_inspection}.

\subsection{Experiment 1}
\label{subsection:experiment_1}

The first experiment covers 500 independent claims randomly selected from patents in 2022 (range: 01/01$\sim$04/12). These patents are new and are not in the USTPO TACD dataset. The performance of different model sizes is shown in Table~\ref{table:evaluation1}. In the table, the first column is the size of the model. The second column is the ``AE ratio,'' and the 456M model is the best (56.8\%). It is noted that all the models perform relatively well, except for the smallest model. Even the 128M model has an AE ratio of more than 50\%. Also, the largest model is not better, although its size is about 13 times larger than the best model. This finding implies that, in a specific domain and with a specific metric, a modest range of model sizes may suffice. Continuing to pursue larger model sizes might not be necessary.

\begin{table*}[]
\caption{PatentGPT-J Performance Evaluation \#1}
\label{table:evaluation1}
\centering
  \begin{tabular}{C{1.6cm} C{2.0cm} C{2.5cm} C{2.5cm} C{2.3cm} C{1.6cm} C{2.0cm}}
    \hline
    \multicolumn{7}{c}{Evaluation by Total Number of Keystrokes} \\ \hline
    \multirow{2}{*}{size} & \multirow{2}{*}{AE ratio ($\uparrow$)} & total w/ & top 10 ($\downarrow$) & out of top 10 ($\downarrow$) & top 1 ($\uparrow$) & total w/o  \\
    & & autocomplete ($\downarrow$) & (auto\slash manual) & (manual) & (auto) & autocomplete \\ \hline\hline
    6B & 56.2\% & 281,789 & \textbf{167,246} & 114,543 & 72,457 & 643,646 \\ \hline
    1.6B & 56.7\% & 278,948 & 168,635 & 110,313 & 73,295 & 643,646 \\ \hline
    \textbf{456M} & \textbf{56.8\%} & \textbf{277,800} & 168,112 & \textbf{109,688} & \textbf{73,600} & 643,646 \\ \hline
    279M & 56.2\% & 281,619 & 169,082 & 112,537 & 72,498 & 643,646 \\ \hline
    191M & 55.6\% & 285,843 & 169,478 & 116,365 & 71,676 & 643,646 \\ \hline
    128M & 50.1\% & 320,907 & 172,604 & 148,303 & 64,284 & 643,646 \\ \hline
    115M & 36.0\% & 411,884 & 177,304 & 234,580 & 44,660 & 643,646 \\ \hline\hline
    \multicolumn{7}{l}{* ``keys w/ autocomplete'' = keystrokes of ``top 10'' + keystrokes of ``out of top 10''} \\ 
    \multicolumn{7}{l}{** AE ratio = (``key w/o autocomplete'' - ``key /w autocomplete'') / ``key w/o autocomplete''} 

  \end{tabular}
\end{table*}

The third column of the table is the total number of keystrokes needed to compose the 500 independent claims with an autocomplete function. The number is the minimal number after taking advantage of the predictions of the model. The lower the number of keystrokes (the higher the AE ratio), the better. The 456M model has the lowest number in this experiment. The last column is the total number of keystrokes without using any autocomplete function. The number is the number of characters that a user has to manually type for all token text. This number is the same for all rows in the table because manual efforts are the same. As a whole, the AE ratio is calculated on the basis of the third column and the last column, as shown in Equation~\ref{eq:1}. The fourth column is the number of keystrokes based on the top 10 predictions of the model. By definition, \emph{``top n''} includes tokens with ranks ranging from \emph{1} to \emph{n}, and \emph{``out of top n''} includes tokens with ranks ranging from \emph{n+1} to \emph{the size of vocabulary} (50,400 in both GPT-J-6B and PatentGPT-J). In summary, the ``top 10'' include all tokens with ranks from 1 to 10. The top 1 is equivalent to rank 1. The details of mixing autocomplete and manual typing in the fourth column are described in section~\ref{subsection:metric} and are omitted here. The fifth column shows the number of keystrokes if a user manually types when the rank of the target token is greater than 10. The relation of the third, fourth, and fifth columns is described below:
  
\begin{equation}
\label{eq:2}
\begin{aligned}
& \text{\footnotesize{total keystrokes with autocomplete = }} \\
& \text{\footnotesize{(keystrokes of ``top 10'') + (keystrokes of ``out of top 10'')}} 
\end{aligned}
\end{equation}

For both ``top 10'' and ``out of top 10, having fewer keystrokes is better. The sixth column (top 1) is a supplementary column showing the top-1 counts in the fourth column (top 10). The higher the counts in the top 1, the better. In this experiment, the 456M model performs the best. This means that using the 456M model, a user can save more keystrokes than using other models. 

\begin{figure*}[ht]
  {\includegraphics[width=\textwidth, keepaspectratio]{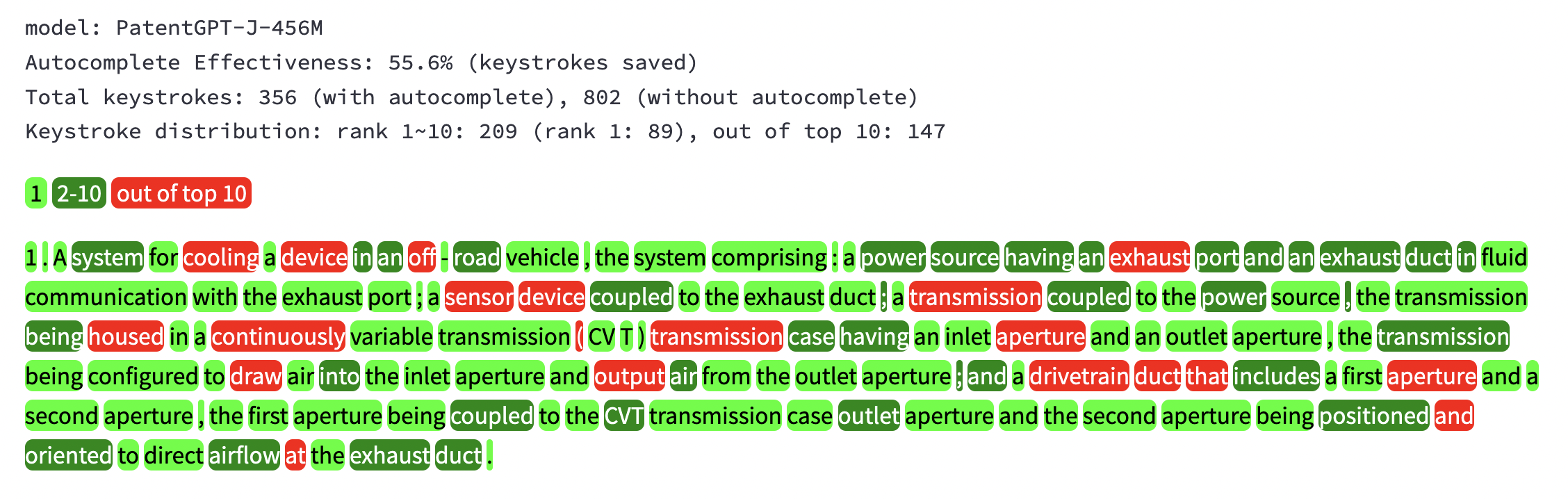}}
  \caption{Inspecting AE Ratio and Keystroke Distribution of Patent 11,248,512 (claim 1).}
  \label{patent_11248512_claim_1}
\end{figure*}

In addition to the performance evaluation in Table~\ref{table:evaluation1}, a visualization tool is developed to inspect the distribution of the ranks on a token basis given a patent claim. All inspection results in this experiment are archived and released through an interactive web page at~\cite{demo1}. The following is one of the results. Fig.~\ref{patent_11248512_claim_1} shows claim 1 of patent 11,248,512 with color highlights and statistics. The highlights are colored in three categories: (1) ``rank 1'' in light green, (2) ``rank 2$\sim$10'' in dark green, and (3) ``out of top 10'' in red. In this style, it is convenient to recognize which tokens can be completed simply by pressing a ``tab'' key (rank 1), or pressing a ``tab'' key and ``downarrow'' key (rank 2$\sim$10) or typing manually if the keystrokes are fewer, or typing all the token text manually (out of top 10). It is also convenient to compare different models in such a visualization. The model selected in this example is the 456M model. The AE ratio is 55.3\%. It should be noted that such a visualization might be a tool for saliency analysis. The visualization is like a saliency map. Salient tokens in the patent context might mean something that is not obvious to a generalized patent language model. Nonobviousness is a legal requirement for patentability in patent law. Nonobviousness is critical not only in patent prosecution but also in patent litigation. The legal implications with respect to patent law will be briefly described in Section~\ref{section:implications}.   

\subsection{Experiment 2}
\label{subsection:experiment_2}
\begin{figure*}[ht]
  {\includegraphics[width=\textwidth, keepaspectratio]{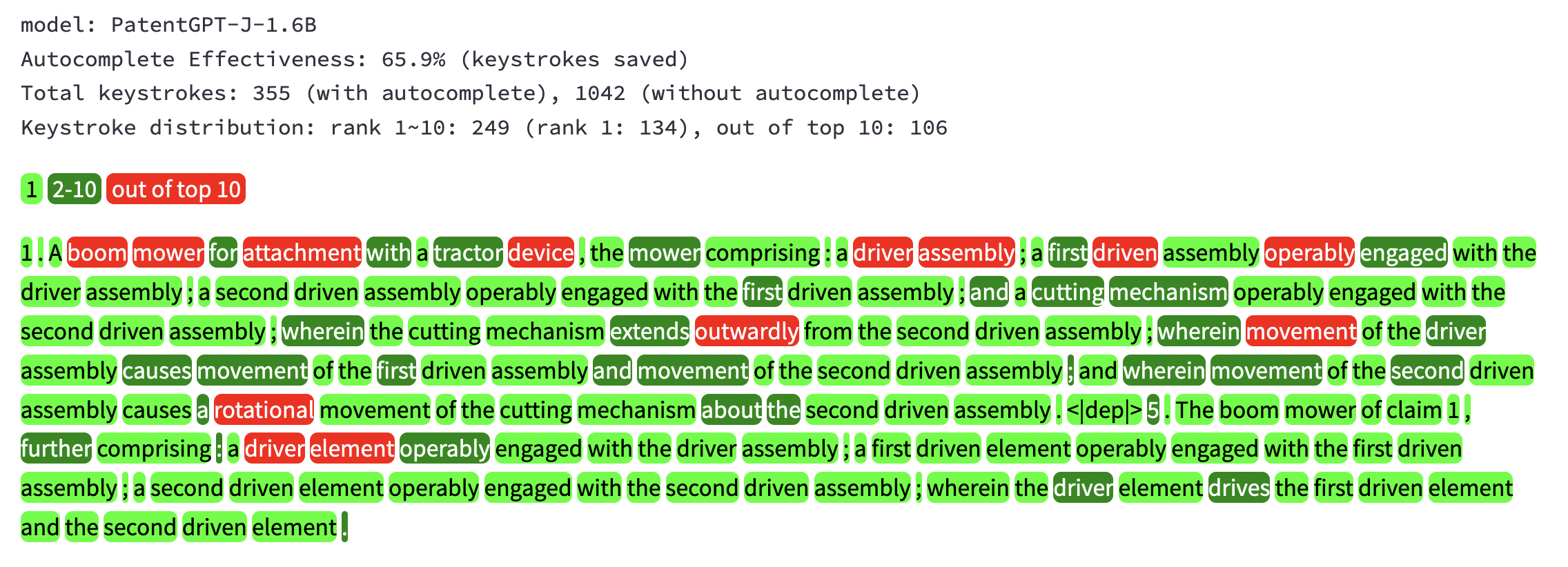}}
  \caption{Inspecting AE Ratio and Keystroke Distribution of Patent 11,212,959 (claim 5).}
  \label{patent_11212959_claim_5}
\end{figure*}

The second experiment covers another 500 patent claims, including independent and dependent claims. These patents are randomly selected from the patents in the first week of 2022 and outside the USTPO TACD dataset. The purpose of this experiment is to verify the findings in the first experiment with different test data. The performance of different models is shown in Table~\ref{table:evaluation2}. The table structure is the same as the previous one. In this experiment, the 1.6B model is the best model in terms of the AE ratio. The 456M model is the second best by a narrow margin. The largest model (6B) is not better even though its size is 3.75 times larger than the 1.6B model. Therefore, it is validated again that continuing to pursue larger model sizes might be unnecessary in a specific domain that has a specific metric such as the AE ratio. A reasonable range of model sizes can be sufficient. The rest of Table~\ref{table:evaluation2} is self-explanatory and the details are omitted for brevity. 
Fig.~\ref{patent_11212959_claim_5} shows the same visualization to inspect the distribution of the ranks on a token basis given a patent claim. The patent claim to analyze is claim 5 of patent 11,212,959. The color codes and their purposes are the same and are omitted here. After comparing different models, the best model for this patent claim is the 1.6B model, as shown below. The AE ratio is 65.9\%.

\begin{table*}[]
\caption{PatentGPT-J Performance Evaluation \#2}
\label{table:evaluation2}
\centering
  \begin{tabular}{C{1.6cm} C{2.0cm} C{2.5cm} C{2.5cm} C{2.3cm} C{1.6cm} C{2.0cm}}
    \hline
    \multicolumn{7}{c}{Evaluation by Total Number of Keystrokes} \\ \hline
    \multirow{2}{*}{size} & \multirow{2}{*}{AE ratio ($\uparrow$)} & total w/ & top 10 ($\downarrow$) & out of top 10 ($\downarrow$) & top 1 ($\uparrow$) & total w/o  \\
    & & autocomplete ($\downarrow$) & (auto\slash manual) & (manual) & (auto) & autocomplete \\ \hline\hline
    6B & 54.9\% & 238,845 & \textbf{150,101} & 88,744 & 73,282 & 529,900 \\ \hline
    \textbf{1.6B} & \textbf{55.3\%} & \textbf{237,015} & 150,423 & \textbf{86,592} & \textbf{73,708} & 529,900 \\ \hline
    456M & 55.2\% & 237,650 & 151,140 & 86,510 & 73,687 & 529,900  \\ \hline
    279M & 54.4\% & 241,708 & 151,275 & 90,433 & 72,687 & 529,900 \\ \hline
    191M & 53.6\% & 245,713 & 151,375 & 94,338 & 72,274 & 529,900 \\ \hline
    128M & 48.5\% & 272,636 & 153,575 & 119,061 & 65,492 & 529,900 \\ \hline
    115M & 35.8\% & 340,012 & 157,193 & 182,819 & 48,702 & 529,900 \\ \hline\hline
    \multicolumn{7}{l}{* ``keys w/ autocomplete'' = keystrokes of ``top 10'' + keystrokes of ``out of top 10''} \\ 
    \multicolumn{7}{l}{** AE ratio = (``key w/o autocomplete'' - ``key /w autocomplete'') / ``key w/o autocomplete''}     
  \end{tabular}
\end{table*}

\subsection{Experiment 3}
\label{subsection:experiment_3}

This experiment calculates the AE ratio of PatentGPT-J models by using non-patent text. The AE ratio should drop because the models were pre-trained with patent corpus only. The purpose of this experiment is to know how much the AE ratio will drop. In this experiment, the non-patent text is the sample ``unicorn'' text generated by the original GPT-2 model and published by OpenAI~\cite{openai_gpt2_blog}, as shown below. 

\lstset{
  basicstyle=\fontsize{7}{9}\selectfont\ttfamily,
  breaklines=true,  
  columns=fullflexible,
  breakindent=0pt
}

\begin{lstlisting}
In a shocking finding, scientist discovered a herd of unicorns living in a remote, previously unexplored valley, in the Andes Mountains. Even more surprising to the researchers was the fact that the unicorns spoke perfect English. The scientist named the population, after their distinctive horn, Ovid's Unicorn. These four-horned, silver-white unicorns were previously unknown to science. Now, after almost two centuries, the mystery of what sparked this odd phenomenon is finally solved.
\end{lstlisting}

\begin{figure*}[ht]
  {\includegraphics[width=\textwidth, keepaspectratio]{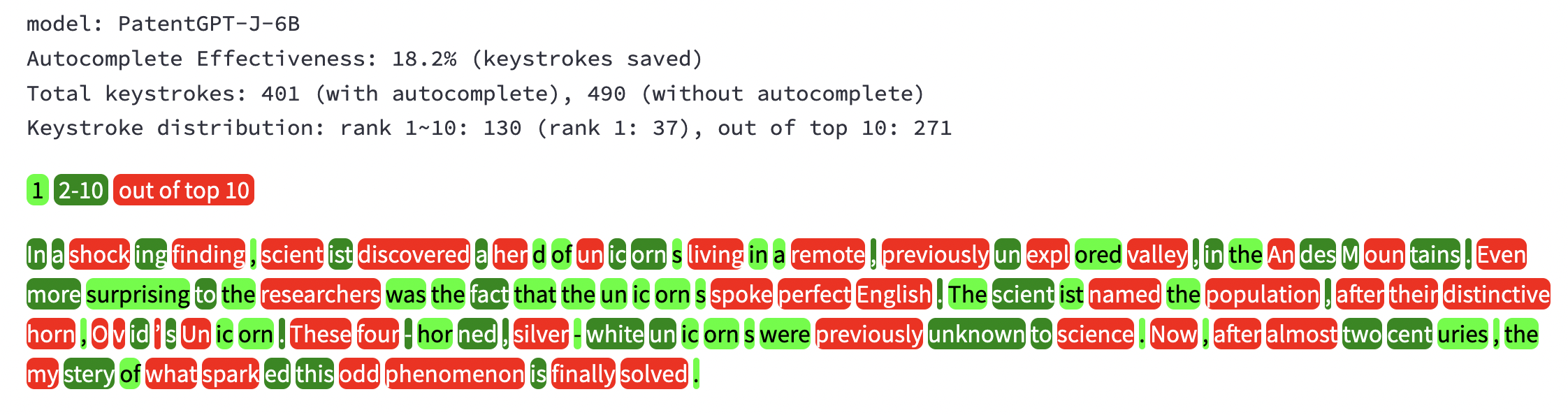}}
  \caption{Inspecting AE Ratio and Keystroke Distribution of the unicorn text.}
  \label{experiment_unicorn2}
\end{figure*}

The performance of different models for non-patent text is shown in Table~\ref{table:evaluation3}. The table structure is the same as the previous ones. In this experiment, the 6B model is the best model in terms of the AE ratio. However, its performance drops significantly to 18.2\%. This ratio shows that an autocomplete function based on PatentGPT-J models is less helpful for composing non-patent text. It is reasonable to see that a domain-specific model fails to predict well out-of-domain content. The rest of Table~\ref{table:evaluation3} is self-explanatory and is omitted.
Fig.~\ref{experiment_unicorn2} shows a similar visualization for inspecting the distribution of the ranks on a token basis given the non-patent text. The model is the 6B model in this figure. It is noted that more tokens are displayed in red (out of top 10) and few are displayed in light green (top 1). Obviously the non-patent text is harder for PatentGPT-J models to predict. Even the largest PatentGPT-J model has a low AE ratio. For all model inspections in this experiment, a web demo is available at~\cite{demo3}. 

\begin{table*}[]
\caption{PatentGPT-J Performance Evaluation \#3}
\label{table:evaluation3}
\centering
  \begin{tabular}{C{1.6cm} C{2.0cm} C{2.5cm} C{2.5cm} C{2.3cm} C{1.6cm} C{2.0cm}}
    \hline
    \multicolumn{7}{c}{Evaluation by Total Number of Keystrokes} \\ \hline
    \multirow{2}{*}{size} & \multirow{2}{*}{AE ratio ($\uparrow$)} & total w/ & top 10 ($\downarrow$) & out of top 10 ($\downarrow$) & top 1 ($\uparrow$) & total w/o  \\
    & & autocomplete ($\downarrow$) & (auto\slash manual) & (manual) & (auto) & autocomplete \\ \hline\hline
    \textbf{6B} & \textbf{18.2\%} & \textbf{401} & 130 & \textbf{271} & \textbf{37} & 490 \\ \hline
    1.6B & 16.7\% & 408 & 125 & 283 & 35 & 490 \\ \hline
    456M & 16.7\% & 408 & 100 & 308 & 29 & 490 \\ \hline
    279M & 14.1\% & 421 & 97 & 324 & 31 & 490 \\ \hline
    191M & 13.9\% & 422 & 105 & 317 & 29 & 490 \\ \hline
    128M & 10.8\% & 437 & 90 & 347 & 25 & 490 \\ \hline
    115M & 8.8\% & 447 & \textbf{86} & 361 & 17 & 490 \\ \hline\hline
    \multicolumn{7}{l}{* ``keys w/ autocomplete'' = keystrokes of ``top 10'' + keystrokes of ``out of top 10''} \\ 
    \multicolumn{7}{l}{** AE ratio = (``key w/o autocomplete'' - ``key /w autocomplete'') / ``key w/o autocomplete''}  \\    
  \end{tabular}
\end{table*}

\subsection{Experiment 4}
\label{subsection:experiment_4}

Apart from testing the AE ratio, this experiment tests the capability of PatentGPT-J models in few-shot learning. According to~\cite{nlpcloud-few-shot-learning}, models like GPT-J, GPT-Neo~\cite{pile}, and GPT-3 are so big that they can easily adapt to many contexts without being re-trained. Giving only a few examples to the model can help it dramatically increase its accuracy. The webpage at~\cite{nlpcloud-few-shot-learning} provides one example of sentiment analysis with GPT-J. Using the example, this experiment tests whether the PatentGPT-J models can correctly predict sentiment in few-shot learning. The non-patent prompt for few-shot learning is as follows:

%\begin{verbatim}[fontsize=\small]
%[language=Python, caption=Python example]
%\begin{lstlisting}[basicstyle=\ttfamily\small]
\lstset{
  basicstyle=\fontsize{7}{9}\selectfont\ttfamily,
}

\begin{lstlisting}
prompt = """Message: Support has been terrible for 2 weeks...
Sentiment: Negative
###
Message: I love your API, it is simple and so fast!
Sentiment: Positive
###
Message: It is really bad! How could it be possible? 
Sentiment: Negative
###
Message: The API is great. It is really good!
Sentiment: Positive
###
Message: GPT-J has been released 2 months ago.
Sentiment: Neutral
###
Message: Your team has been amazing, thanks!
Sentiment:"""
\end{lstlisting}
%\end{verbatim}

Table~\ref{table:sentiment_analysis} shows the results of the experiment. For each model, the test is run 1,000 times with the same prompt and parameters (\emph{top\_p=0.9} and \emph{temperature=0.75}). Although the smallest model (115M) cannot produce anything sensible, surprisingly, the best model (1.6B) predicts 731 times correctly. It is also surprising that a model as small as the 191M model is already capable of predicting the correct answer 526 times. In terms of the format for possible answers (Positive, Negative, or Neutral), the 191M model can predict the answers in correct format 871 times (=526+85+260). The results of the experiment and the sample code are released as ``Unreasonable effectiveness in Few-Shot Learning'' at~\cite{jiehsheng_PatentGPT-J_github}.

\begin{table}[]
\caption{Sentiment Analysis in Few-Shot Learning}
\label{table:sentiment_analysis}
\centering
  \begin{tabular}{c c c c c}
  %\begin{tabular}{c c c c c}
    \hline
    %\multicolumn{5}{c}{Evaluation by Minimal Number of Keystrokes} \\ \hline
    % ? & ? & ? & ? & ? \\ \hline
    size & positive & negative & neutral & others \\ \hline\hline
    6B & 636 & 263 & 86 & 15 \\ \hline
    \textbf{1.6B} & \textbf{731} & 167 & 67 & 35 \\ \hline
    456M & 535 & 152 & 37 & 276 \\ \hline
    279M & 586 & 285 & 6 & 123 \\ \hline
    \textbf{191M} & \textbf{526} & \textbf{85} & \textbf{260} & 129 \\ \hline
    128M & 358 & 268 & 24 & 350 \\ \hline
    115M & 0 & 0 & 0 & 1,000 \\ \hline\hline
  \end{tabular}
\end{table}

\subsection{Statistics}
\label{subsection:statistics}

\begin{figure}[ht]
  {\includegraphics[width=0.5\textwidth, keepaspectratio]{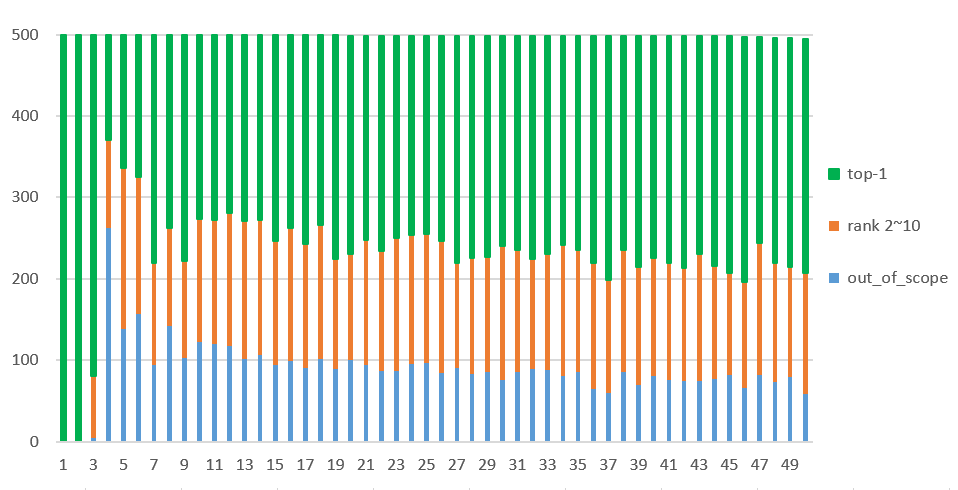}}
  \caption{Statistics of First 50 Token Positions (PatentGPT-J-456M) (x: token position, y: counts)}
  \label{456M_statistics_50_token_positions}
\end{figure}

\begin{figure}[ht]
  {\includegraphics[width=0.5\textwidth, keepaspectratio]{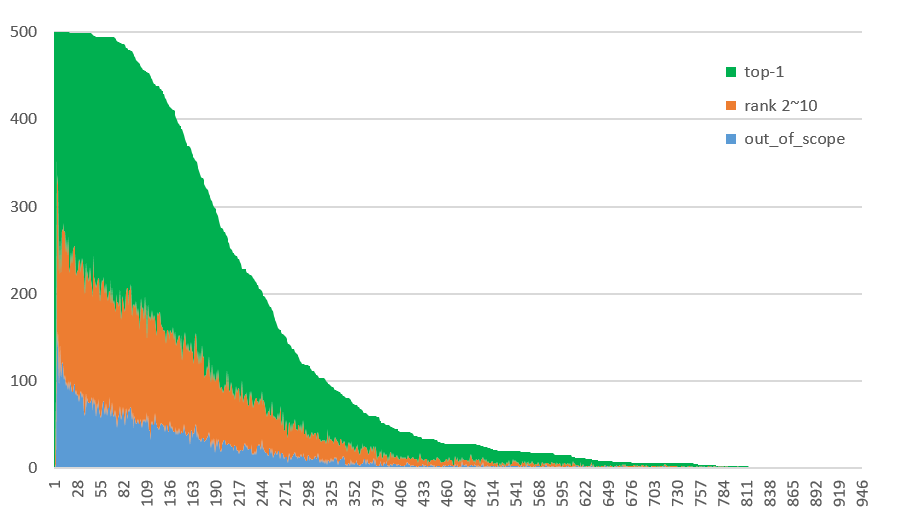}}
  \caption{Statistics of All Token Positions (PatentGPT-J-456M) (x: token position, y: counts)}
  \label{456M_statistics_all_token_positions}
\end{figure}

In experiment~\ref{subsection:experiment_1}, the best model is the 456M model. In this section, based on the model, two charts are provided to show the distribution of token ranks at each token position. Fig.~\ref{456M_statistics_50_token_positions} shows the distribution of ranks for the first 50 tokens of the 500 patent claims in the experiment. In this figure, the percentage of the ``top 1'' counts increases gradually along the token position. The first three tokens could be ignored because a patent claim always begins with ``1. A'' or ``1. An'' in terms of formality. To observe the finding with more tokens, Fig.~\ref{456M_statistics_all_token_positions} shows the distribution of ranks with all token sequences of the 500 patent claims in the same experiment. A similar finding is that the percentage of the ``top 1'' counts continues to dominate along the token position. Based on these findings, a PatentGPT-J model may save more keystrokes with an autocomplete function if the patent text to draft is longer.  

\subsection{Interactive Inspection}
\label{subsection:interactive_inspection}

\begin{figure}[ht]
  {
  \includegraphics[width=0.5\textwidth, keepaspectratio]{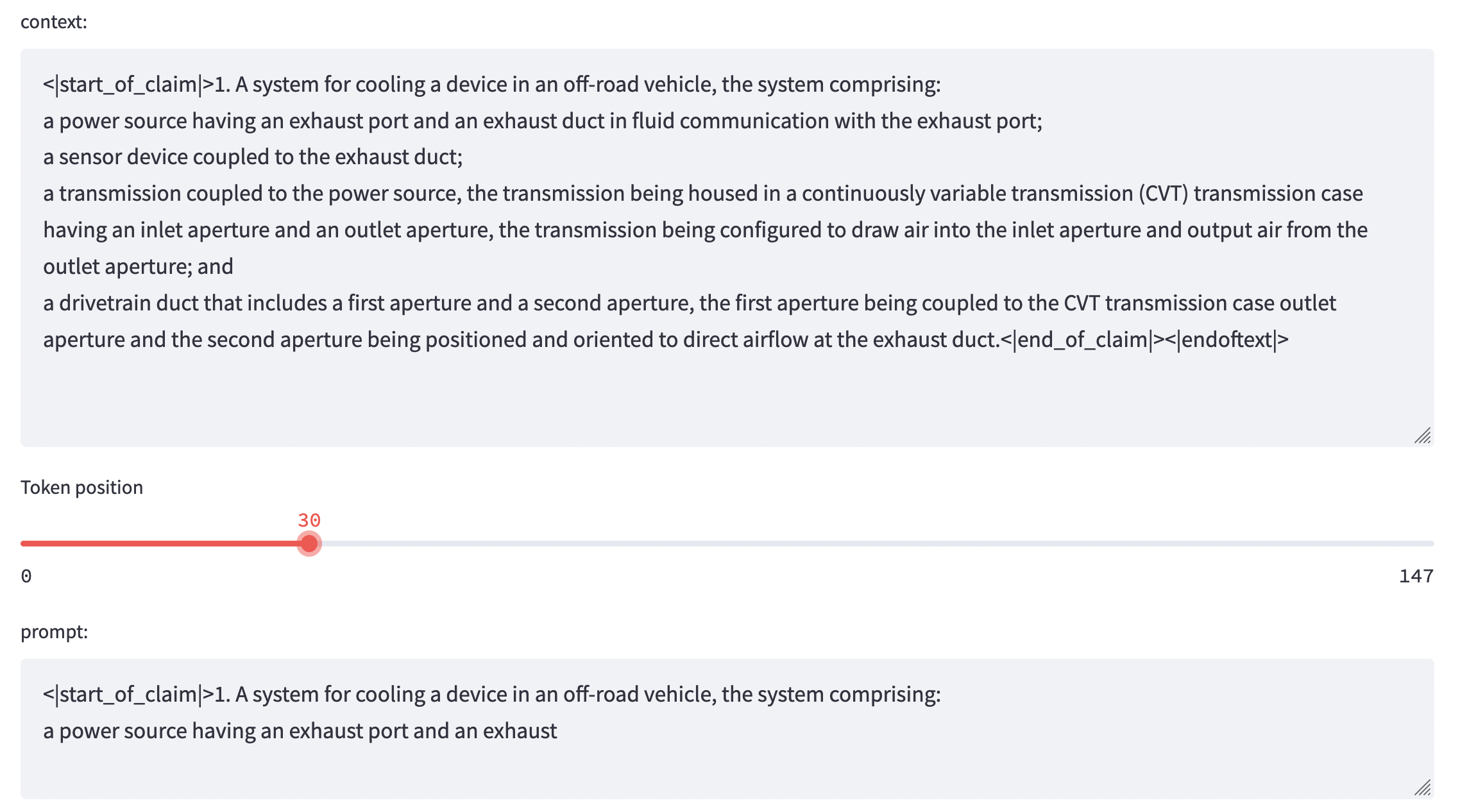}
  \includegraphics[width=0.5\textwidth, keepaspectratio]{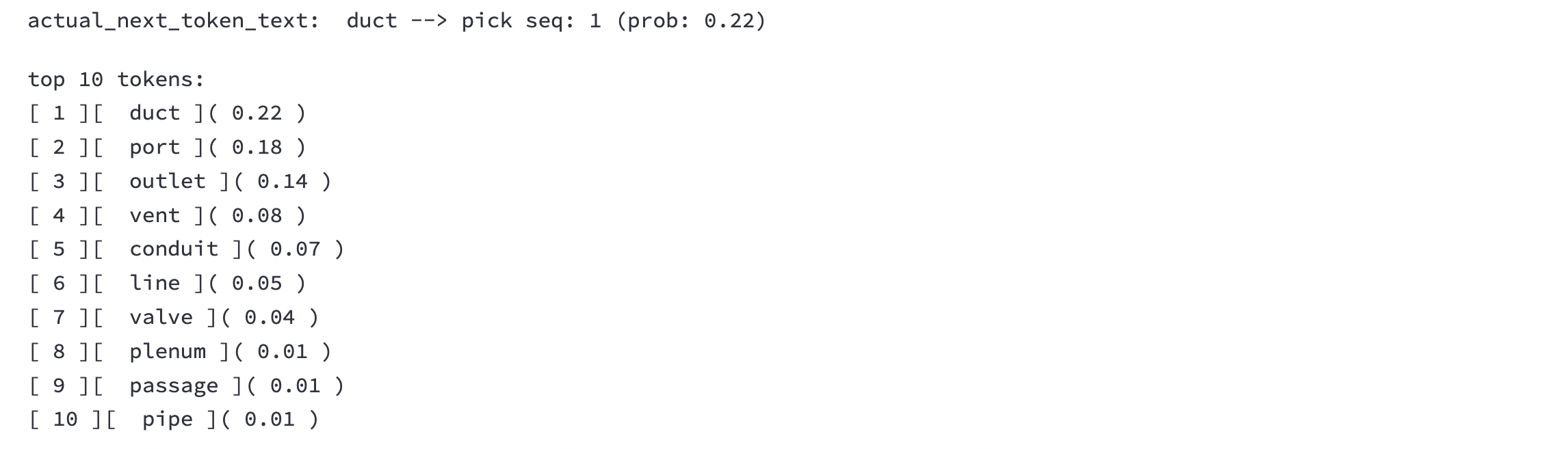}
  \includegraphics[width=0.5\textwidth, keepaspectratio]{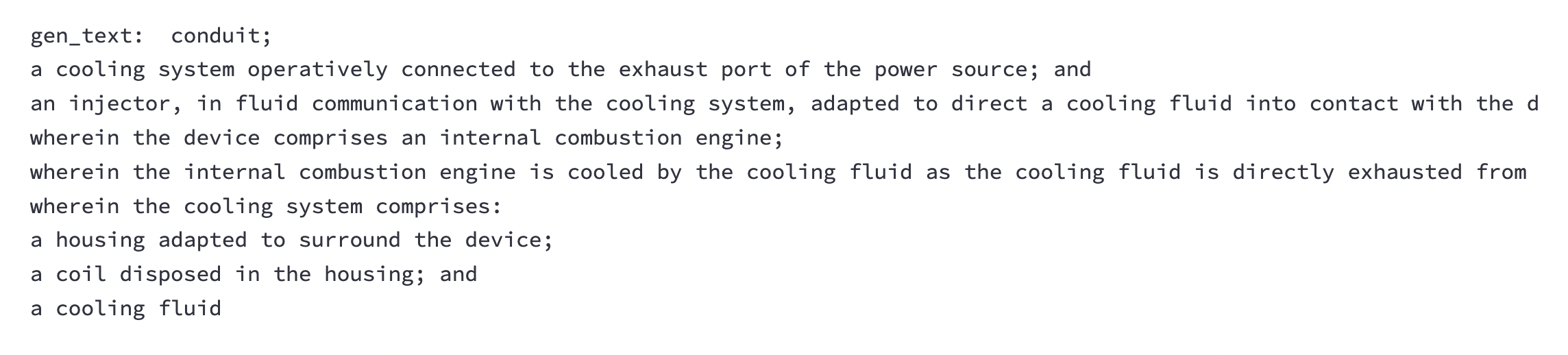}
  }
  \caption{Inspection: (1) next top 10 tokens, (2) generated text at position \emph{n}}
  \label{probe}
\end{figure}

The interactive tool in this section provides a user interface (UI) for inspecting the details of the top 10 predictions at each token position. All evaluations in experiment~\ref{subsection:experiment_2} and their visualization are released as an interactive web page at~\cite{demo2}. The UI is interactive in real time because the details of all evaluation results were generated and stored in advance when calculating the AE ratio. In Algorithm~\ref{alg:counting_keystrokes} (line: 10), the model predicts the top 10 tokens based on the prompt. When implemented, what is stored are not only the predicted top 10 tokens but also the related details at that point of time. The stored details make it possible for the UI to show the results interactively without triggering the time-consuming model inference again.
For example, Fig.~\ref{probe} shows the details at token position 30. Explaining from the top of the UI, the first part ``context'' is a text area showing the entire patent text to calculate the AE ratio. The second part is a slider to adjust the token position. A token position \emph{n} means the \emph{nth} token in the ``context.'' Based on the token position, the third part of the UI shows all the tokens before that position. The text area ``prompt'' is the input for the model to generate the next token. In this example, the next actual token in the ``context'' is the ``duct'' and the token is the top 1 choice in the prediction of the model. Therefore, the rank (pick sequence) is 1. The probability of picking the top 1 choice is 0.22, as shown below the ``prompt'' part. However, whether the model will pick the top 1 choice is a random process. In fact, in this example, the model picks the fifth token (``conduit'') during text generation. In the UI, the next part shows the top 10 tokens and their text and probabilities. The last part of the UI shows the generated text at that token position, as shown in the text area ``gen\_text.'' Depending on the parameters for sampling, the model may generate patent text with different degrees of randomness. The model continues to generate text until it reaches the end of the claim. 
In this example, the model generates ``conduit; a cooling system operatively connected to the exhaust port of the power source; and ...'' (carriage return omitted). For saving time, the length of text generation was limited to 128 tokens in model inference. Although the patent claim generated is not finished in this example, the generated text is coherent, fluent, and sufficient for brainstorming. If not told, a patent professional may not recognize the text as machine-generated immediately. 

A practical use case for inventors based on such a tool might be to explore inventive ideas in the generated patent text. Since the model is specific to the patent domain, the model can generate domain-specific text that may contain ideas different from the prompt. Different ideas may augment inventors to invent better and faster. This idea of augmentation was proposed as the ``Augmented Inventing'' idea in previous research~\cite{jiehsheng03}. The difference and progress here is that, in this manuscript, the patent text is generated at scale and at each token position. All text is generated and stored in one batch. The UI in this section is a tool for an inventor to quickly slide through each position and brainstorm with all the generated patent text.  
If needed, multiple samples of patent text can be generated with the same or different parameters in sampling. How to enhance the tool after conducting field tests and collecting users' feedback is a crucial step to move forward. 

\section{Legal Implications} % and Broader Impacts
\label{section:implications}

This section is philosophical with several open questions for legal scholars. 
Section 103 of the US Patent Law defines ``A patent for a claimed invention may not be obtained, ..., if the \emph{differences} between the claimed invention and the \emph{prior art} are such that the claimed invention as a whole would have been \emph{obvious} before the effective filing date of the claimed invention to \emph{a person having ordinary skill in the art} to which the claimed invention pertains. Patentability shall not be negated by \emph{the manner in which the invention was made} (emphasis added).'' PatentGPT-J models may trigger several legal questions from the perspective of Section 103. 
First, can a PatentGPT-J model be considered as or approximate the person having ordinary skill in the art (PHOSITA)? A PHOSITA is a hypothetical person in patent law. The reason why a neural model may serve the role of PHOSITA is that: The training of a neural model is about generalizing among training data. If the training data are past patents (prior arts), the model is a generalized model based on prior arts. If such a generalized model approximates the ordinary skill in the prior arts, the model may be deemed as the PHOSITA. In this line of thinking, a fundamental question to be explored will be: Can the ``generalization'' in the technical sense be equivalent to the ``ordinary'' in the legal sense? 

Second, can we use a PatentGPT-J model to measure the nonobviousness requirement in the patent law? A PatentGPT-J model is a generative model. Assuming that the model is a PHOSITA, what the model can generate with high probabilities might be deemed obvious to the model. If high probabilities can approximate obviousness, a generative model such as PatentGPT-J may act as a PHOSITA to measure nonobviousness with low probabilities. In this line of thinking, a fundamental question to be debated will be: Can the ``high or low probabilities'' in the technical sense be equivalent to the ``obviousness or nonobviousness'' in the legal sense? Conceptually, it should be reasonable to say that if one thing is harder (of lower probability or predictability) for an inventor to predict, it should be something less obvious to the inventor.

In KSR,~\footnote{KSR Int'l Co. v. Teleflex Inc., 550 U.S. 398 (2007)} the Supreme Court particularly emphasized ``the need for caution in granting a patent based on the combination of elements found in the prior art,'' and importantly, the Supreme Court reaffirmed principles based on its precedent that ``[t]he combination of familiar elements according to known methods is likely to be obvious when it does no more than predictable results.'' In this manuscript, if the tokens of a language model can be considered the elements of the patent language, can we say that the combination of high-probability tokens is the combination of familiar elements? Can we say that the combination is likely to be obvious when the language model yields no more than predictable results (tokens)? In this line of thinking, a fundamental question to be debated will be: Can the ``tokens'' in the technical sense be equivalent to the ``elements'' in the legal sense?

Third, can an inventor combine the patent text generated from a PatentGPT model and make the entire patent application eligible for patent protection? According to Section 103 of the patent law, patentability shall not be negated by \emph{the manner in which the invention was made}. Although the statute was codified to overturn the flash-of-genuine doctrine, can we say that the manner of using a patent language model to conceive new ideas or combine generated patent text will not negate any patentability? 

\section{Conclusion \& Looking Forward}
\label{section:conclusion}

In this manuscript, several generative language models in the patent domain are pre-trained from scratch and released. To evaluate model performance, a human-centric metric is proposed to measure the ratio of keystrokes that can be saved by autocompletion based on the generative model. A higher ratio means that more keystrokes are saved. The proposed metric is a post hoc analysis. Given a granted patent claim, the metric can be utilized to show how close the model may generate the granted patent claim. Although the largest model in this manuscript is state-of-the-art in terms of model size, it is found that the largest model is not necessarily the best for the human-centric metric. For inspecting the results of the experiments in this manuscript, several visualization tools are also provided. The evaluation of metric and model in this manuscript focuses only on patent claims. Looking forward, the same metric can be applied to the entire patent document to measure the effectiveness of autocompletion and generative models. A model capable of being generative can be a tool for being creative. 
The importance of generative language models is their potential in the future to facilitate creativity and innovation in the patent domain.

\break
\break

\noindent\textbf{Acknowledgements}
\break
%\section{Acknowledgments}
The research reported in this manuscript has been funded by the
Ministry of Science and Technology (MOST) in Taiwan (Project ID:
111-2222-E-A49-005). In addition, the author would like to thank TensorFlow Research Cloud (TRC) greatly for providing powerful computing resources. 

\printcredits

\bibliographystyle{model1-num-names}
%\bibliographystyle{cas-model2-names}

\bibliography{citation}

\begin{thebibliography}{23}
\expandafter\ifx\csname natexlab\endcsname\relax\def\natexlab#1{#1}\fi
\providecommand{\url}[1]{\texttt{#1}}
\providecommand{\href}[2]{#2}
\providecommand{\path}[1]{#1}
\providecommand{\DOIprefix}{doi:}
\providecommand{\ArXivprefix}{arXiv:}
\providecommand{\URLprefix}{URL: }
\providecommand{\Pubmedprefix}{pmid:}
\providecommand{\doi}[1]{\href{http://dx.doi.org/#1}{\path{#1}}}
\providecommand{\Pubmed}[1]{\href{pmid:#1}{\path{#1}}}
\providecommand{\bibinfo}[2]{#2}
\ifx\xfnm\relax \def\xfnm[#1]{\unskip,\space#1}\fi
%Type = Misc
\bibitem[{Wang and Komatsuzaki(2021)}]{gpt-j-github}
\bibinfo{author}{B.~Wang}, \bibinfo{author}{A.~Komatsuzaki},
  \bibinfo{title}{{GPT-J-6B: A 6 Billion Parameter Autoregressive Language
  Model}},
  \bibinfo{howpublished}{\url{https://github.com/kingoflolz/mesh-transformer-jax}},
  \bibinfo{year}{2021}.
%Type = Inproceedings
\bibitem[{Brown et~al.(2020)Brown, Mann, Ryder, Subbiah, Kaplan, Dhariwal,
  Neelakantan, Shyam, Sastry, Askell, Agarwal, Herbert-Voss, Krueger, Henighan,
  Child, Ramesh, Ziegler, Wu, Winter, Hesse, Chen, Sigler, Litwin, Gray, Chess,
  Clark, Berner, McCandlish, Radford, Sutskever, and
  Amodei}]{GPT-3-NEURIPS2020}
\bibinfo{author}{T.~Brown}, \bibinfo{author}{B.~Mann},
  \bibinfo{author}{N.~Ryder}, \bibinfo{author}{M.~Subbiah},
  \bibinfo{author}{J.~D. Kaplan}, \bibinfo{author}{P.~Dhariwal},
  \bibinfo{author}{A.~Neelakantan}, \bibinfo{author}{P.~Shyam},
  \bibinfo{author}{G.~Sastry}, \bibinfo{author}{A.~Askell},
  \bibinfo{author}{S.~Agarwal}, \bibinfo{author}{A.~Herbert-Voss},
  \bibinfo{author}{G.~Krueger}, \bibinfo{author}{T.~Henighan},
  \bibinfo{author}{R.~Child}, \bibinfo{author}{A.~Ramesh},
  \bibinfo{author}{D.~Ziegler}, \bibinfo{author}{J.~Wu},
  \bibinfo{author}{C.~Winter}, \bibinfo{author}{C.~Hesse},
  \bibinfo{author}{M.~Chen}, \bibinfo{author}{E.~Sigler},
  \bibinfo{author}{M.~Litwin}, \bibinfo{author}{S.~Gray},
  \bibinfo{author}{B.~Chess}, \bibinfo{author}{J.~Clark},
  \bibinfo{author}{C.~Berner}, \bibinfo{author}{S.~McCandlish},
  \bibinfo{author}{A.~Radford}, \bibinfo{author}{I.~Sutskever},
  \bibinfo{author}{D.~Amodei},
\newblock \bibinfo{title}{Language models are few-shot learners},
\newblock in: \bibinfo{editor}{H.~Larochelle}, \bibinfo{editor}{M.~Ranzato},
  \bibinfo{editor}{R.~Hadsell}, \bibinfo{editor}{M.~Balcan},
  \bibinfo{editor}{H.~Lin} (Eds.), \bibinfo{booktitle}{Advances in Neural
  Information Processing Systems}, volume~\bibinfo{volume}{33},
  \bibinfo{publisher}{Curran Associates, Inc.}, \bibinfo{year}{2020}, pp.
  \bibinfo{pages}{1877--1901}. \URLprefix
  \url{https://proceedings.neurips.cc/paper/2020/file/1457c0d6bfcb4967418bfb8ac142f64a-Paper.pdf}.
%Type = Article
\bibitem[{Lee and Hsiang(2020)}]{jiehsheng03}
\bibinfo{author}{J.-S. Lee}, \bibinfo{author}{J.~Hsiang},
\newblock \bibinfo{title}{Patent claim generation by fine-tuning openai gpt-2},
\newblock \bibinfo{journal}{World Patent Information} \bibinfo{volume}{62}
  (\bibinfo{year}{2020}) \bibinfo{pages}{101983}.
%Type = Inproceedings
\bibitem[{Lee(2020)}]{jiehsheng05}
\bibinfo{author}{J.-S. Lee},
\newblock \bibinfo{title}{Controlling patent text generation by structural
  metadata},
\newblock in: \bibinfo{booktitle}{Proceedings of the 29th ACM International
  Conference on Information and Knowledge Management}, CIKM '20,
  \bibinfo{publisher}{Association for Computing Machinery},
  \bibinfo{address}{New York, NY, USA}, \bibinfo{year}{2020}, p.
  \bibinfo{pages}{3241–3244}. \URLprefix
  \url{https://doi.org/10.1145/3340531.3418503}.
  \DOIprefix\doi{10.1145/3340531.3418503}.
%Type = Inproceedings
\bibitem[{Lee and Hsiang(2021)}]{jiehsheng09}
\bibinfo{author}{J.-S. Lee}, \bibinfo{author}{J.~Hsiang},
\newblock \bibinfo{title}{Prior art search and reranking for generated patent
  text},
\newblock in: \bibinfo{booktitle}{Proceedings of the 2nd Workshop on Patent
  Text Mining and Semantic Technologies (PatentSemTech) 2021, co-located with
  the 44th International ACM SIGIR Conference on Research and Development in
  Information Retrieval (SIGIR 2021)}, PatentSemTech 2021,
  \bibinfo{year}{2021}, pp. \bibinfo{pages}{18--24}. \URLprefix
  \url{http://ceur-ws.org/Vol-2909/paper2.pdf}.
%Type = Article
\bibitem[{Gao et~al.(2020)Gao, Biderman, Black, Golding, Hoppe, Foster, Phang,
  He, Thite, Nabeshima, Presser, and Leahy}]{pile}
\bibinfo{author}{L.~Gao}, \bibinfo{author}{S.~Biderman},
  \bibinfo{author}{S.~Black}, \bibinfo{author}{L.~Golding},
  \bibinfo{author}{T.~Hoppe}, \bibinfo{author}{C.~Foster},
  \bibinfo{author}{J.~Phang}, \bibinfo{author}{H.~He},
  \bibinfo{author}{A.~Thite}, \bibinfo{author}{N.~Nabeshima},
  \bibinfo{author}{S.~Presser}, \bibinfo{author}{C.~Leahy},
\newblock \bibinfo{title}{The {P}ile: An 800gb dataset of diverse text for
  language modeling},
\newblock \bibinfo{journal}{arXiv preprint arXiv:2101.00027}
  (\bibinfo{year}{2020}).
%Type = Misc
\bibitem[{USPTO(2022)}]{uspto_bulkdata}
\bibinfo{author}{USPTO}, \bibinfo{title}{Bulk data storage system},
  \bibinfo{howpublished}{\url{https://bulkdata.uspto.gov/}},
  \bibinfo{year}{2022}.
%Type = Misc
\bibitem[{Komatsuzaki(2021)}]{gpt-j-6b-blog}
\bibinfo{author}{A.~Komatsuzaki}, \bibinfo{title}{{GPT-J-6B: 6B JAX-Based
  Transformer}},
  \bibinfo{howpublished}{\url{https://arankomatsuzaki.wordpress.com/2021/06/04/gpt-j/}},
  \bibinfo{year}{2021}.
%Type = Misc
\bibitem[{Wang et~al.(2021)Wang, Biderman, and de~la
  Rosa}]{gpt-j-how-to-fine-tune}
\bibinfo{author}{B.~Wang}, \bibinfo{author}{S.~Biderman},
  \bibinfo{author}{J.~de~la Rosa}, \bibinfo{title}{{How to Fine-Tune GPT-J}},
  \bibinfo{howpublished}{\url{https://github.com/kingoflolz/mesh-transformer-jax/blob/master/howto_finetune.md}},
  \bibinfo{year}{2021}.
%Type = Misc
\bibitem[{Radrof et~al.(2018)Radrof, Wu, Child, Luan, Amodei, and
  Sutskever}]{gpt2_Radrof01}
\bibinfo{author}{A.~Radrof}, \bibinfo{author}{J.~Wu},
  \bibinfo{author}{R.~Child}, \bibinfo{author}{D.~Luan},
  \bibinfo{author}{D.~Amodei}, \bibinfo{author}{I.~Sutskever},
  \bibinfo{title}{Language models are unsupervised multitask learners},
  \bibinfo{year}{2018}.
%Type = Misc
\bibitem[{Izadi et~al.(2022)Izadi, Gismondi, and Gousios}]{CodeFill}
\bibinfo{author}{M.~Izadi}, \bibinfo{author}{R.~Gismondi},
  \bibinfo{author}{G.~Gousios}, \bibinfo{title}{Codefill: Multi-token code
  completion by jointly learning from structure and naming sequences},
  \bibinfo{year}{2022}. \URLprefix \url{https://arxiv.org/abs/2202.06689}.
%Type = Inproceedings
\bibitem[{Aye et~al.(2021)Aye, Kim, and Li}]{Learning_Autocompletion}
\bibinfo{author}{G.~A. Aye}, \bibinfo{author}{S.~Kim}, \bibinfo{author}{H.~Li},
\newblock \bibinfo{title}{Learning autocompletion from real-world datasets},
\newblock in: \bibinfo{booktitle}{2021 IEEE/ACM 43rd International Conference
  on Software Engineering: Software Engineering in Practice (ICSE-SEIP)},
  \bibinfo{year}{2021}, pp. \bibinfo{pages}{131--139}.
  \DOIprefix\doi{10.1109/ICSE-SEIP52600.2021.00022}.
%Type = Misc
\bibitem[{EleutherAI(2020)}]{EleutherAI_pile_uspto}
\bibinfo{author}{EleutherAI}, \bibinfo{title}{pile-uspto},
  \bibinfo{howpublished}{\url{https://github.com/EleutherAI/pile-uspto}},
  \bibinfo{year}{2020}.
%Type = Misc
\bibitem[{cfoster0(2020)}]{cfoster0_uspto-patent-data-parser}
\bibinfo{author}{cfoster0}, \bibinfo{title}{uspto-patent-data-parser (forked
  from tamerkhraisha)},
  \bibinfo{howpublished}{\url{https://github.com/cfoster0/uspto_patent_data_parser}},
  \bibinfo{year}{2020}.
%Type = Misc
\bibitem[{TamerKhraisha(2020)}]{TamerKhraisha_uspto-patent-data-parser}
\bibinfo{author}{TamerKhraisha}, \bibinfo{title}{uspto-patent-data-parser},
  \bibinfo{howpublished}{\url{https://github.com/TamerKhraisha/uspto-patent-data-parser}},
  \bibinfo{year}{2020}.
%Type = Misc
\bibitem[{Lee(2022)}]{PatentGPT-J-training_loss}
\bibinfo{author}{J.-S. Lee}, \bibinfo{title}{{Training loss of
  PatentGPT-J-models}},
  \bibinfo{howpublished}{\url{https://wandb.ai/jason-lee/patent-gpt-j-d-4096/reports/PatentGPT-J-models--VmlldzoyMDgwNjI5?accessToken=u30ezzk5mxvfjdgqsyhhajfgm1mnoi9plj4x832jafpu4zl82zr5fc4fk3bn9zyl}},
  \bibinfo{year}{2022}.
%Type = Misc
\bibitem[{Wang(2021)}]{GPT-J-training_loss}
\bibinfo{author}{B.~Wang}, \bibinfo{title}{{Training loss of GPT-J-6B}},
  \bibinfo{howpublished}{\url{https://wandb.ai/eleutherai/mesh-transformer-jax/reports/GPT-J-Pretraining--VmlldzoxMjMzOTgy}},
  \bibinfo{year}{2021}.
%Type = Misc
\bibitem[{Lee(2022{\natexlab{a}})}]{jiehsheng_PatentGPT-J_github}
\bibinfo{author}{J.-S. Lee}, \bibinfo{title}{Patent{GPT-J} repository},
  \bibinfo{howpublished}{\url{https://github.com/jiehsheng/PatentGPT-J}},
  \bibinfo{year}{2022}{\natexlab{a}}.
%Type = Misc
\bibitem[{Lee(2022{\natexlab{b}})}]{demo1}
\bibinfo{author}{J.-S. Lee}, \bibinfo{title}{{(Demo 1) Inspecting AE ratio and
  keystroke distribution}},
  \bibinfo{howpublished}{\url{https://huggingface.co/spaces/patent/demo1}},
  \bibinfo{year}{2022}{\natexlab{b}}.
%Type = Misc
\bibitem[{OpenAI(2019)}]{openai_gpt2_blog}
\bibinfo{author}{OpenAI}, \bibinfo{title}{{Better Language Models and Their
  Implications}},
  \bibinfo{howpublished}{\url{https://openai.com/blog/better-language-models}},
  \bibinfo{year}{2019}.
%Type = Misc
\bibitem[{Lee(2022)}]{demo3}
\bibinfo{author}{J.-S. Lee}, \bibinfo{title}{{(Demo 3) Inspecting AE Ratio and
  Keystroke Distribution of the unicorn text.}},
  \bibinfo{howpublished}{\url{https://huggingface.co/spaces/patent/demo3}},
  \bibinfo{year}{2022}.
%Type = Misc
\bibitem[{Cloud(2022)}]{nlpcloud-few-shot-learning}
\bibinfo{author}{N.~Cloud}, \bibinfo{title}{How to use gpt-3, gpt-j and
  gpt-neo, with few-shot learning},
  \bibinfo{howpublished}{\url{https://nlpcloud.io/effectively-using-gpt-j-gpt-neo-gpt-3-alternatives-few-shot-learning.html}},
  \bibinfo{year}{2022}.
%Type = Misc
\bibitem[{Lee(2022)}]{demo2}
\bibinfo{author}{J.-S. Lee}, \bibinfo{title}{{(Demo 2) Inspection: (1) next top
  10 tokens, (2) generated text at position \emph{n}}},
  \bibinfo{howpublished}{\url{https://huggingface.co/spaces/patent/demo2}},
  \bibinfo{year}{2022}.

\end{thebibliography}

\bio{}
\endbio


\end{document}